\documentclass[10pt]{llncs}
\usepackage{rotating}
\usepackage{pdflscape}
\usepackage{amsmath}
\usepackage{amsfonts}
\usepackage{amssymb}
\usepackage{amstext}
\usepackage{latexsym,stmaryrd}
\usepackage[ruled, linesnumbered,noend]{algorithm2e}

\usepackage{inferance}
\usepackage{deduc}
\def\infty{\text{BUG}}

\renewcommand{\leq}{\mathrel{\leqslant}}

\renewcommand\ruleformat[1]{\ #1}

\begin{document}
\title{Revisiting the Learned Clauses Database Reduction Strategies}
\titlerunning{Learned Clauses Database Reduction}
\author{Said Jabbour and Jerry Lonlac and Lakhdar Sais and Yakoub Salhi}

\institute{CRIL - CNRS, Universit\'e d'Artois, Lens, France\\
\{jabbour, lonlac, sais, salhi\}@cril.fr}

\maketitle

\begin{abstract}
In this paper, we revisit an important issue of CDCL-based SAT solvers, namely the learned clauses database management policies.  
Our motivation takes its source from a simple observation on the remarkable performances of both random and size-bounded reduction strategies.   
We first derive a simple reduction strategy, called Size-Bounded Randomized strategy (in short SBR), that combines maintaing short clauses (of size bounded by $k$), while deleting randomly clauses of size greater than $k$.  The resulting strategy outperform the state-of-the-art, namely the LBD based one, on SAT instances taken from the last SAT competition.
 Reinforced by the interest of keeping short clauses, we propose several new dynamic variants, and we discuss their performances.  

%In this paper, we revisit an important issue of CDCL-based SAT solvers, namely the learned clauses database management policy.  
%Our motivation has its source from a simple observation on the remarquable performances of both random and size-bounded reductions strategies.   
%We first derive a simple reduction strategy, called Size-Bounded Randomized strategy (in short SBR), that combines maintaing short clauses (of size bounded by $k$), while deleting randomly clauses of size greater than $k$. Reinforced by the interest of keeping short clauses, we propose a new static/dynamic measure that allows us to quantify the relevance of a given learned clause w.r.t. the current state of the search process. Intuitively,    a short learned clause with literals assigned most often at the top of the current branch of the search tree is considered relevant. Using this measure, we derive a second reduction strategy, called Relevance-Based Strategy (in short RBS), that allows us to keep learned clauses that are more likely to cut branches at the top of the search tree. 
%The implementation of both strategies (SBR and RBS) at the top of MiniSAT 2.2 outperform the state-of-the-art SAT solvers, on application instances taken from the last SAT competition. 
\end{abstract}
 
\section{Introduction}
%The SAT problem, i.e., the problem of checking whether a Boolean formula in conjunctive normal form (CNF) is satisfiable or not, is central to many domains in computer science and artificial intelligence including constraint satisfaction problems (CSP), automated planning, non-monotonic reasoning, VLSI correctness checking, etc.  
Today, SAT has gained a considerable audience with the advent of a new generation of solvers able to solve large instances encoding real-world problems. 
%and the demonstration that these solvers represent important low-level building blocks for many important fields, e.g., SAT modulo theory, Theorem proving, Model checking, Quantified boolean formulas, Maximum Satisfiability, Pseudo boolean,  etc. 
These solvers, often called \emph{modern SAT solvers} \cite{Moskewicz01,MiniSat03}, are based on the classical DPLL procedure \cite{Davis62} enhanced with: (i) an efficient implementation of unit propagation through incremental and lazy data structures,  (ii) restart policies \cite{Gomes1998,kautz02dynamic}, (iii) activity-based variable selection heuristics (VSIDS-like) \cite{Moskewicz01}, and (iv) {\it clause learning} \cite{Marques-Silva96,Moskewicz01}.   Clause learning  is now recognized as one of the most important component of Modern SAT solvers. The main idea is that when a current branch of the search tree leads to a conflict, clause learning aims to derive a clause that succinctly  expresses the causes of the conflict. Such learned clause is then used to prune the search space. Clause learning also known in the literature as Conflict Driven Clause Learning (CDCL) refers now to the most known and used First UIP learning scheme, first integrated in the SAT solver Grasp \cite{Marques-Silva96} and efficiently implemented in zChaff \cite{Moskewicz01}. Most of the SAT solvers, integrate this strong learning scheme. Theoretically, by integrating clause learning to DPLL-like procedures \cite{Davis62}, the obtained SAT solver formulated as a proof system is as powerful as general resolution \cite{PipatsrisawatD09,Adnan10}. 

In practice, the efficiency of CDCL-based SAT solvers heavily depends on the strategy used to manage the learned clauses database. Indeed, as at each conflict, a new clause is added to the learned clauses database, its size grows exponentially. To avoid this combinatorial explosion, several learned clauses database management strategies have been proposed (e.g. \cite{Moskewicz01,MiniSat03,AudemardS09,AudemardLMS11,GJS-13-1}). These strategies aims to maintain a learned clause database of reasonable size by eliminating clauses deemed irrelevant to the subsequent search. All these management strategies follow a predefined cleaning time-sequence where the interval between two successive reduction steps is more or less important. At each conflict, an activity is associated to the learned clause (static strategy). Such heuristic-based activity aims to weight each clause according to its relevance to the search process. In the case of dynamic strategies, such clauses activities are dynamically updated. The reduction of the learned clauses database consists in eliminating inactive or irrelevant clauses.  Even if all the proposed learned clause deletion strategies are shown to be empirically efficient, determining the most relevant clause to the search process remains a challenging task.  It is important to note that the efficiency of most of the state-of-the-art learned clauses  management strategies heavily depends on the cleaning frequency and on the amount of clauses to be deleted each time. 

Different implementations of SAT solvers are proposed each year to the SAT competition, they include several enhancements to the main components of CDCL-based solvers. The SAT competitions, races and challenges stimulate the development of efficient implementations. However, such a race to the most efficient implementation has led to increasingly complex solvers with several number of parameters. 
Such parameters are either static (fixed before search), or dynamic, their values are conditionally set during search by observing the search behavior. These sophisticated implementations 
increase the difficulty to understand what is essential from what is not. In \cite{JoaoSak11}, an empirical analysis focusing on the principal techniques that have contributed to the impressive performance of modern SAT solvers has been conducted. This is really a first step towards a deep understanding of modern SAT solvers. 
%To be convinced, it suffices to look inside, the different state-of-the-art SAT solvers. Our general goal, is develop the most simple but efficient SAT solver, by keeping only the most important parts and enhancements made on the current modern SAT solvers. In our opinion, such study must start by trying to quantify the contribution of each component to the practical success of SAT solvers. For example, how many problems we loose if the restart is not used, or the learned clauses database is not reduced, or chronological backtracking is inhibited and so on.

In this paper, we revisit an important issue of CDCL-based SAT solvers, namely the learned clauses database management policy.   It is {\it important to note} at this point that considering short clauses as the most relevant ones (size-bounded based reduction strategies) have been proposed  since 1996 by Marques Silva and Sakallah (Grasp \cite{Marques-Silva96}) and Bayardo and Schrag (RelSAT \cite{Bayardo97}).  Most of the state-of-the-art SAT solvers maintain systematically only binary clauses, while for clauses of size greater that two, several sophisticated relevance based measures have been proposed to predict the most relevant ones. 

Our motivation in this work has its source from a simple observation on the remarkable performances of size-bounded reduction strategies. From this first  results, we decided to investigate other "naive" strategies such as random and FIFO (First In First Out). The goal is to quantify the performance gap between these simple strategies and those implemented in the state-of-the-art SAT solvers.  
Secondly, we derive a simple reduction strategy, called Size-Bounded Randomized Strategy (in short SBR), that combines maintaing short clauses (of size bounded by $k$), while deleting randomly clauses of size greater than $k$. Reinforced by the interest of keeping short clauses, we propose several  new dynamic measures that allow us to quantify the relevance of a given learned clause w.r.t. the current state of the search process. Intuitively,  {\it a short learned clause with literals assigned most often at the top of the current branch of the search tree is considered more relevant}. Several relevance-based strategies are then derived, allowing us to keep learned clauses that are more likely to cut branches at the top of the search tree.  All these strategies are integrated at the top of MiniSAT 2.2 and compared with the state-of-the-art SAT solvers on application instances taken from the last SAT competition.  To confirm the superiority of size-based measures against LBD-based ones, we also present the results obtained by substituting LBD with dynamic size-based activity inside $Glucose$ $3.0$. The obtained results bring up to date old strategies, such as size-bounded proposed more than fifteenth years ago \cite{Marques-Silva96,Bayardo96acomplexity,Bayardo97} (see Section \ref{sec:relwork} - Related Works). These results are not surprising, since short clauses are usually preferred for their ability to reduce further the search space. This is also why all the SAT solvers keep the learned binary clauses.  

%depends on addIf we look at the learned clauses database reduction component, in addition to the measure used to quantify the quality of a given clause, several other sophisticated ingredient are added to manage the amount of clauses to be deleted, the frequency of such deletion process and so on.  

%Our goal is to first consider the performance of a random deletion strategy against the ones used in some state-of-the-art satisfiability solvers. We then propose a new reduction strategy based on a very simple and more intuitive measure that allow us to predict the most relevant clauses. The intuitive idea is based on the following simple fact "a clause is better when its literals are involved at the top of search tree". 

The paper is organized as follows. We first recall related works in Section~\ref{sec:relwork}. Then, after some technical background and preliminary definitions (Section~\ref{sec:tbpf}), we motivate our study by investigating the size-bounded, random and FIFO  deletion strategies in Section \ref{sec:size-random}. In Section \ref{sec:sbr}, we present our first reduction strategy, called Size-Bounded Randomized strategy. In Section \ref{sec:rbs}, we describe several relevant based deletion strategies, that allow us to keep learned clauses that are more likely to cut branches at the top of the search tree. All these new strategies are integrated at the top of MiniSAT 2.2 and compared to the state-of-the-art SAT solvers $Glucose$  $3.0$ and $Lingeling$ $2013$. We also present the results obtained by integrating size-bounded based strategies inside $Glucose$ $3.0$. A discussion is provided in Section \ref{sec:discuss}, before concluding.

 \section{Related Works}
\label{sec:relwork}
%Preventing combinatorial explosion by maintaining a relevant set of clauses is not novel issue. 
%Several previous research works addressing this issues, include deletion-directed search in resolution-based proof procedures \cite{Gelperin:73}, relevance or size bounded learning in CSP-look-back techniques \cite{Bayardo97}, etc. 
In this section, we give a non exhaustive review of some important deletion strategies as described by the authors. To our knowledge, maintaing a relevant set of clauses goes back to the developments of efficient resolution-based proof procedures in automated theorem proving.  In \cite{Gelperin:1973}, D. Gelperin proposed different  strategies that attempt to determine the satisfiability of a set of input clauses while at the same time minimizing the cardinality of the set of retained clauses. 

In constraint satisfaction and propositional satisfiability problems, to overcome the overhead of unrestricted learning, several strategies have been proposed by Bayardo et al. \cite{Bayardo97,Bayardo96acomplexity}, including size-bounded and relevance-bounded learnings. They defined size-bounded (respectively, relevance-bounded) learning of order $i$, which retains indefinitely only clauses derived reasons containing $i$ or fewer variables (respectively, maintains any reason that contains at most $i$ variables whose assignments have changed since the reason was derived).  In SAT, this issue is first investigated in GRASP by Joao Marques Silva et al. \cite{Marques-Silva96}. Indeed, in order to guarantee the worst case growth of the clause database to be polynomial in the number of variables, in \cite{Marques-Silva96} the authors propose a selective strategy on the clauses that have to be added to the clause database. More precisely, given an integer parameter $k$, conflict-induced clauses whose size (number of literals) is less than $k$ are marked green (added to the database) while those of size greater than k are marked red and kept around only while they are unit clauses, i.e., a red clause is deleted when it contains more than one free (unassigned) literal.  

Like many other solvers, Chaff~\cite{Moskewicz01}  supports the deletion of added conflict clauses to avoid a memory explosion. Essentially, Chaff uses scheduled lazy clause deletion. When each clause is added, it is analyzed to determine at what point in the future, if any, the clause should be deleted. The metric used is relevance, such that when more than $n$ (where $n$ is typically 100-200) literals in the clause will become unassigned for the first time, the clause will be marked as deleted. The actual memory associated with deleted clauses is recovered with an infrequent monolithic database compaction step. 

In Berkmin \cite{Goldberg20071549}, the authors consider that more recently deduced clauses are more valuable because it took more time to deduce them from the original set of clauses. The learned clause database is considered as a queue (First In First Out). The Berkmin deletion strategy maintains short clauses (size less than 8) combined with queue representation of the learned clauses database.    

MiniSat \cite{MiniSat03} aggressively deletes learned clauses based on an activity heuristic similar to the one for variables. A learnt clause is considered as irrelevant if its activity or its involvement in recent conflict analysis is marginal \cite{MiniSat03}. The limit on how many learned clauses are allowed is increased after each restart. This strategy has been improved in MiniSAT 2.2. 
%Keeping the number of clauses low seems to be particulary important for some small but hard problems. The actual activity heuristic currently used is admittedly a weak point of MiniSat. 

More recently, Audemard and Simon \cite{AudemardS09}, use the number of different levels (LBD - Literal Block Distance) involved in a given learnt clause to quantify the quality of learnt clauses. The set of different levels present in a learned clause has been used in \cite{euipTR} to prove the optimality of the First UIP scheme.  In \cite{AudemardS09}, clauses with smaller LBD are considered as more relevant. The LBD measure is integrated in MiniSAT solver leading to Glucose, one of the state-of-the-art SAT solvers.  LBD based measure is also exploited in Lingeling \cite{Lingeling12}, SAT13 \cite{knuthSAT13w} the CDCL-based SAT solver designed by D. Knuth, and some other SAT solvers entering the last SAT competition 2013 (e.g. $gluebit\_clasp$ $1.0$, $BreakIDGlucose$ $1$, $glueminisat$ $2.2.7j$).  Another dynamic management policy of the learnt clauses database is proposed in \cite{AudemardLMS11}. It is based on a dynamic freezing and activation
principle of the learnt clauses. At a given search state, using a relevant selection function based on progress saving \cite{darwiche07}, it activates the most promising learned clauses while freezing irrelevant
ones. 

In \cite{GJS-13-1},  we proposed two new measures to predict the quality of learned clauses. The first one is based on the backtracking level (BTL), while the second is based on a notion of distance, defined as the difference between the maximum and minimum assignment levels of the literals involved in the learned clause. 

Finally, size-bounded clause sharing strategies are also considered in several portfolio and divide and conquer based parallel SAT solvers (e.g. ManySAT \cite{ManySATJSAT} and PMSat \cite{LPL08}).  
 
  %strategies  resolution based proof procedures is  
%Using In this paper, we first review some of the state-of-the-art learned clauses database management strategies. 

%Strategies agressive (Adnan).... 

% ICI

%%%%%%%%%%%%
\section{Technical Background and Preliminary Definitions}
\label{sec:tbpf}
A propositional formula ${\cal F} $ in {\it Conjunctive Normal Form (CNF)} is a
conjunction of {\it clauses}, where a clause is 
a disjunction of {\it literals}. 
A literal is a positive ($x$) or negated ($\neg{x}$) 
propositional variable.  The two literals $x$ and $\neg{x}$ are called {\it complementary}. 
 Let $c$ be a clause, $|c|$ denotes the size of $c$ (its number of literals). 
% A CNF formula can also be seen as a set of clauses, and a clause as a set of literals. 
%Let us recall that any Boolean formula can be translated to CNF using linear Tseitin encoding \cite{Tseitin68}. 
A {\it unit} clause is a clause containing only one literal (called {\it unit literal}), while a  binary clause contains exactly two literals. An \emph{empty clause}, denoted $\perp$, is interpreted as  false (unsatisfiable), whereas an \emph{empty CNF formula}, denoted $\top$, is interpreted as true (satisfiable). The set of variables occurring in ${\cal F}$ is denoted $V_ {\cal F}$. An {\it interpretation} $\rho$ of a propositional  formula ${\cal F}$ is a function which  associates a value $\rho(x)\in \{false, true\}$ to some of the variables $x \in V_ {\cal F}$. 
$\rho$ is \emph{complete} if it assigns a value to every $x \in V_ {\cal F}$, and \emph{partial} otherwise.
An interpretation $\rho$ is alternatively represented by a set of literals, i.e., $\rho = \bigcup_{x\in V_{\cal F}} f(x)$, 
where $f(x)=x$ (respectively $f(x)= \neg x$), if $\rho(x)= true$ (respectively $\rho(x) = false$). 
A {\it model} of a formula ${\cal F}$ is an  interpretation $\rho$ that  satisfies the  formula.

%We consider the  conjunctive normal form (CNF) representation for the propositional formulas. A {\it CNF formula}  $\Phi$  is a
%conjunction of {clauses}, where a {\it clause} is a disjunction of {literals}. 
%A {\it literal} is a positive ($x$) or negated ($\neg{x}$) 
%propositional variable.  The two literals $x$ and $\neg{x}$ are called {\it complementary}. 
%
%A CNF formula can also be seen as a set of clauses, and a clause as a set of literals. 
%%Let us recall that any propositional formula can be translated to CNF using linear Tseitin's encoding \cite{Tseitin68}. 
%We denote by $Var(\Phi)$ the set of propositional variables occurring in $\Phi$.\\
%~\\
%An {\it interpretation} ${\cal B}$ of a propositional formula $\Phi$ is a function which  associates a value ${\cal B}(x )\in\{0, 1\}$ ($0$ corresponds to $false$ and $1$ to $true$)
%to the variables $x \in Var(\Phi)$.  A {\it model} of a formula $\Phi$ is an  interpretation ${\cal B}$ that  satisfies the  formula.  {\it SAT problem} consists in deciding if a given  formula 
%admits a model or not.  \\
%~\\
%We denote by $\bar{x}$  the complementary literal of $x$, i.e., if  $x = p$ then $\bar{x}=\neg p$ and if  $x = \neg p$ then $\bar{x}=p$.
%For a set of literals $L$, $\bar{L}$ is defined as $\{\bar{l} ~|~ l \in L\}$. 
%Moreover, $\overline{ \cal B}$ (${\cal B}$ is an interpretation over $Var(\Phi)$) corresponds to the clause $\bigvee_{p\in Var(\Phi)} f(p )$, 
%where if  ${\cal B}(p )=0$ then $f(p )=p$,  otherwise $f( p)=\neg p$. 

Let us now introduce some notations and terminology on SAT solvers
based on the Davis Logemann Loveland procedure, commonly called DPLL \cite{Davis62}. 
DPLL  is a backtrack search procedure;
at each node the assigned literals (decision literal and the propagated ones) are labeled with the
same \emph{decision level} starting from 1 and increased at each branching.
The current decision level is the highest decision level  in the assignment stack. When a conflict is encountered 
After backtracking, some variables are unassigned, and the current decision level is decreased accordingly. 
At level $i$, the current partial assignment $\rho$ can be represented as a sequence of 
decision-propagation of the form
$\langle( x_k^i),x_{k_1}^i, x_{k_2}^i,\dots, x_{k_{n_k}}^i \rangle$ where the
first literal $x_k^i$ corresponds to  the decision literal $x_k$ assigned at level $i$ 
and  $x_{k_j}^i$ for $1\leq j\leq n_k$  represent (unit) literals propagated at level $i$.  Such a partial interpretation (sequence of decision-propagations) associated to a given node of the search tree is called a {\it partial ordered interpretation}. Let $x\in\rho$, we denote by $lev(x, \rho)$ the assignment level of $x$ in $\rho$. Let $\rho$ be a partial interpretation, and $c$ a clause, we define $c^i$ as the projection of $c$ to the literals of $c$ assigned at level $i$, i.e., $c^i=\{x| lev(x, \rho)=i\}$. This set is called a block in \cite{AudemardS09}.  

A  conflict driven clause learning (CDCL) SAT solver explore the search space by making successively a sequence of decision/propagation. When a conflict is encountered a conflict clause is derived by resolution  on the clauses involved in the unit propagation process (encoded as an implication graph). Such learned conflict clause is then added to the learned clauses database. It allows us to produce a unit (asserting) literal at an earlier level. Then, the CDCL-based SAT solver backtracks to that level, propagates the asserting literal and repeats the sequence of decision/propagation until a model is found or an empty clause is derived. The search regularly restarts, and the learned clauses database is regularly reduced by eliminating irrelevant clauses. These different components are interrelated. For example, the restart have strong effect on the learning component \cite{Huang07,Biere08Restart,PipatsrisawatD09Restart,PipatsrisawatD09,ManySATJSAT,AudemardS12}. Some of the state-of-the-art SAT solvers such as $Glucose$ use aggressive restart, very helpful for solving unsatisfiable SAT instances. 
%In this paper, we revisit the learned clause database deletion strategies. 

%For a given level $i$, we define $\rho^i$ (respectively $c^i$) as the projection of $\rho$ (respectively $c$) to literals assigned at a level $i$. 

%A CDCL-based SAT solver can be seen as {\it variant of the DPLL procedure} sketched above, with the following main components:
%\begin{enumerate}
%\item {\it Activity-Based Heuristics}:  
%\item {\it Conflict Driven Clause Learning}:
%\item {\it Restarts}:
%\item {\it Learned Clauses Database Management}:
%\end{enumerate}
%In this paper, we deals with the learned clauses database management component. 
%=====================================================================
\section{Size, Random and FIFO Based Deletion Strategies }
\label{sec:size-random}

As mentioned in the introduction, our main aim is to first quantify the performance gap 
between the state of the art learned clause deletion strategies and some basic strategies including random deletion based one. 
In this section, we illustrate the performance of size, random and FIFO deletion strategies. All these static strategies are integrated without any other modification to MiniSAT 2.2. The three new versions of MiniSAT  are obtained as follows:
\begin{itemize}
\item $Size$-$MiniSAT$: at each conflict, the activity of the learned clause $c$ is set to $|c|$, i.e,. ${\cal A}(c) = |c|$ 
\item $Rand$-$MiniSAT$: at each conflict, the activity of the learned clause $c$ is set to random real value $w\in [0..1]$, i.e., ${\cal A}(c) = drand(random\_seed)$. We used exactly the random function of MiniSAT with the same random\_seed to allow reproducible results. 
\item {$FIFO$-$MiniSAT$}: In this version, the learned clauses are managed using a queue. Each time a reduction is performed, the oldest clauses are deleted. 
\end{itemize}

%To obtain the two versions $Size$-$MiniSAT$ and $Rand$-$MiniSAT$, each time a learned clause $c$ is derived, we set its activity ${\cal A}(c)$  to $|c|$ and  $rand()$ (a real value $w\in [0..1]$) respectively. 

% and $nbConflicts$ respectively. 
%rand(random\_seed).  Le meme solver donne la meme chose deux fois le seed as in minisat, la meme fonction de generation que minisat. 

%1/4 de la base et les 3/4 random. trie par la taille on prend le 1/4,  l'autre 1/4, rand(seed) et on prend 1/4 supplémentaire. 
%and {$FIFO$-$MiniSAT$}

%\begin{itemize}
%\item {$Size$-$MiniSAT$}: each time a learned clause is generated, its activity is set to its size: ${\cal A}(c) = |c|$). 
%\item {$Rand$-$MiniSAT$}: each time a learned clause is generated, its activity is set to a real random value ${\cal A}(c)\in [0..1]$ : ${\cal A}(c) = rand()$.
%\item {$FIFO$-$MiniSAT$}: each time a learned clause is derived, its activity is set to the curent number of conflicts : ${\cal A}(c) = $nbConflicts.
%\end{itemize}

\begin{table}[h]
\begin{small}
\centering
\begin{tabular}{|l|c|c|}
 \hline
{$Solvers$}  & {\#Solved (\#SAT - \#UNSAT)} & {Average Time}  \\ \hline
{$MiniSAT$ } &{201 (122 - 79)} & {956.78s}    \\  \hline
{$Glucose$ }& {216 (104 - 112)} & {\bf 807.62s} \\  \hline
{$Lingeling$ }& {{\bf 233} (119 - \bf 114)} & {1090.99s} \\  \hline
\hline 
{$Size$-$MiniSAT$ }  & {220 ({\bf 126} - 94)} & {1226.27s} \\  \hline
{$Rand$-$MiniSAT$}& {191 (121 - 70)} & {1071.05s} \\  \hline
{$FIFO$-$MiniSAT$ }  &{173 (119 - 54)} & 862.50s    \\  \hline
%{$Minisat 2.2 + Weight$} & { {\bf 234 (125} - 109)} & {1132.08s}  \\  \hline
\end{tabular}
\vspace{0.2cm}
\caption{A Comparative Evaluation of Size/Rand/FIFO-MiniSAT}
\label{tab:CompCriteres}
\end{small}
\end{table}
For all the deletion strategies defined in this paper,  clauses with smaller activities are considered more relevant. For the deletion frequency and the amount of deleted clauses, we follow the same strategy implemented in MiniSAT 2.2.   For all the experiments presented in this paper, we run the SAT solvers on the 300 application instances taken from SAT competition 2013. All the instances are preprocessed by SatElite\cite{satelite} before running the SAT solver. We used Intel Xeon quad-core machines with 32GB of RAM running at 2.66 Ghz. For each instance, we used a timeout of $5000$ seconds of CPU time.

In Table \ref{tab:CompCriteres}, we give the comparative experimental evaluation of our three version of $MiniSAT$ with $Glucose$ $3.0$ and $Lingeling$ $2013$ the best solvers of the SAT competition 2013 (application category). In the second column, we give the total number of solved instances (\#Solved). We also mention, the number of instances proven satisfiable (\#SAT) and unsatisfiable (\#UNSAT) in parenthesis. The third column show the average CPU time in seconds (total time on solved instances divided by the number of solved instances).

From this first experiment, the state-of-the-art SAT solvers $Lingeling$ $2013$ and $Glucose$ $3.0$ solve 233 and 216 instances respectively. The solver $MiniSAT$ solve 201 instances. Let us comment on the performance of the other versions of MiniSAT without any other enhancement.  The $Size$-$MiniSAT$ solver is able to solve 220 instances (4 more instance than $Glucose$ $3.0$). Looking at the average CPU time, Glucose 3.0 performs better. Other important observation can be drawn from this first experiment. On satisfiable instances $Size$-$MiniSAT$ performs better than all the other solvers. It solves 126 instances. The last remark that can be drawn, $Rand$-$MiniSAT$ solves more satisfiable instances (121) than the state-of-the-art solvers $Glucose$ $3.0$ and $Lingeling$ $2013$.  Finally $FIFO$-$MiniSAT$ is clearly the worst, particularly on unsatisfiable instances (only 54 instances are solved). However, it remains competitive on satisfiable instances and solves 15 instances more than $Glucose$ $3.0$.

\section{Size-Bounded Randomized  Strategy}
\label{sec:sbr}
The remarquable results obtained in Section \ref{sec:size-random} by $Size$-$MiniSAT$, where short clauses are considered relevant is clearly encouraging. In this section, we propose a new simple reduction strategy, called Size-Bounded Randomized strategy (in short SBR), that combines maintaing short clauses (of size bounded by $k$), while deleting randomly clauses of size greater than $k$. One of the main drawback of the clause-size based activity, is that larger clauses are often considered irrelevant. On some hard problems, such drastic selection of learned short clauses might have negative effects. To overcome this problem, we introduced some randomization to the size based approach. More precisely, SBR strategy is defined as follows: given a upper bound $k$ on the length of the learned clauses, each time a learned clause $c$ is derived, if its size is less than $k$ then ${\cal A}(c)= |c|$, otherwise ${\cal A}(c)= k + drand(random\_seed)$. In other words, we still prefer short clauses, while for clauses of size larger than $k$, they are considered of equal size $k$ with an additional random value. In this way, larger clauses can be selected randomly. The solver $SBR(k)$-$MiniSAT$ (where $k$ is the upper bound size) implements the above described strategy. 
%To obtain such implementation, we added only three lines of codes to MiniSAT 2.2.  
To determine the best upper bound size $k$, we run $SBR(k)$-$MiniSAT$ with $k = 2$, $5$, $10$, $12$ and $15$.

\begin{table}[h]
\centering
\begin{small}
\begin{tabular}{l|c|c|}
\hline
% \cline{2-3} & \multicolumn{2}{c|}{Minisat2.2}  \\ \hline
 \multicolumn{1}{|l|}{Solvers}  & {\#Solved (\#SAT - \#UNSAT)} & {Average Time}  \\ \hline
  \multicolumn{1}{|l|}{$Glucose$ $3.0$}& {216 (104 - 112)} & {\bf 807.62s} \\  \hline
 \multicolumn{1}{|l|}{$Lingeling$ $2013$}& {233 ( 119 - \bf 114)} & {1090.99s} \\  \hline
 \hline
 \multicolumn{1}{|l|}{$SBR(2)$-$MiniSAT$}& {196 (122 - 74)} & {1064.17s} \\  \hline
 \multicolumn{1}{|l|}{$SBR(5)$-$MiniSAT$}& {218 (118 - 100)} & {1213.36s} \\  \hline
 \multicolumn{1}{|l|}{$SBR(10)$-$MiniSAT$}& { {231} (124 - 107)} & {1158.25s} \\  \hline
 \multicolumn{1}{|l|}{$SBR(12)$-$MiniSAT$}& { {\bf 236} ({\bf 128} - 108)} & {1266.53s} \\  \hline
 \multicolumn{1}{|l|}{$SBR(15)$-$MiniSAT$}& {228 (120 - 108)} & {1226.96s} \\  \hline
 \hline
%\multicolumn{1}{|l|}{$FIFO$}  &{173 (119 - 54)} & {\bf 862.50s}    \\  \hline
% \multicolumn{1}{|l|}{$Size$}  & {220 ({\bf 126} - 94)} & {1226.27s} \\  \hline
% \multicolumn{1}{|l|}{$CSIDS$} &{201 (122 - 79)} & {956.78s}    \\  \hline
% \multicolumn{1}{|l|}{$LBD$} & {206 ( 119 - 87)} & { {1139.69s}}   \\  \hline
% \multicolumn{1}{|l|}{$Weight$} & {\bf 222 (122 - 100)} & {1157.31s}  \\  \hline
\end{tabular}
\vspace{0.2cm}
\caption{A Comparative Evaluation of $SBR(k)$-$MiniSAT$}
\label{tab:compar1}
\end{small}
\end{table}

The results are depicted in Table \ref{tab:compar1}.  $SBR(12)$-$MiniSAT$ obtains the best performances. It solves $236$ instances, $3$ instance more than $Lingeling$ $2013$ and $20$ instances more than $Glucose$ $3.0$. These remarquable results, bring up to date, old strategies such as size-bounded strategies proposed more than fifteenth years ago \cite{Marques-Silva96,Bayardo96acomplexity,Bayardo97}. The results also show that adding some randomization to allow selecting clauses of larger size allows us to introduce some diversification to the resolution derivation of CDCL-Based SAT solvers. It is important to note that our proposed implementation can be obtained by adding three lines of codes to $MiniSAT$ $2.2$ without any tuning or additional enhancements. It is worth noticing that the solvers  $SBR(k)$-$MiniSAT$,  for $k=5,10,12,15$, outperform the solver $Glucose$ $3.0$.

\section{Towards Relevant-Based Deletion Strategies} 
\label{sec:rbs}
Reinforced by the interest of keeping short clauses, our goal in this section is to show if some other variants of these size-based strategies, can lead to better performance. More precisely, our aim is to design dynamic deletions strategies, where the activities of the learned clauses are updated during search. In most of the state-of-the-art SAT solvers, the activities of the learned clauses are updated dynamically. 

\subsection{Clause Relevance-Based Dynamic Measures}
\label{sec:rbsDyn}
Let us explain the idea that has guided us in the design of these new dynamic variants.  Intuitively, {\it a short learned clause with literals assigned most often at the top of the current branch of the search tree is considered more relevant}. 

Given a partial interpretation $\rho$  and $c$ a clause from the learned clause database.  Our first dynamic size based deletion strategy is defined as follows: the initial activity of $c$ is set to $|c|$. Suppose now that $l\in\rho$ is assigned by unit propagation at level $d$, thanks to the learnt clause $c$. In this case, the new activity of $c$ is set to $d$ if $d < |c|$. This measure allows us to keep learned clauses that are more likely to cut branches at the top of the search tree. For example, for two clauses of equal size, we prefer those containing literals assigned at the top of the search tree. By integrating this measure to MiniSAT 2.2, we obtain our first variant $SizeD$-$MiniSAT$. 

Following the same idea, we propose a second version $Size(k)D$-$MiniSAT$. The goal is to introduce a threshold $k$ on the length of the clauses, in order to update only the activity of clauses greater than $k$, and keeping the activity of short clauses (size less than $k$) static. More precisely, the initial activity of $c$ is set as follows: if $|c|\leq k$ then ${\cal A}(c) = |c|$, otherwise ${\cal A}(c) = k+|c|$. Each time, a learned clause $c$ is the reason of a unit propagated literal at decision level $d$, its activity is updated as follows: if $k+d < {\cal A}(c)$ then ${\cal A}(c) = k+d$. In this way, the activity of a clause with size less than $k$ is not updated. 

The third variant is defined as follows. Let $\rho$ be a partial interpretation leading to a conflict at decision level $d$ and $c$ its associated learned clause. The activity of $c$ is initially set to ${\cal A}(c) = \sum_{i=1}^{d} i \times |c^i|$. During search, each time $c$ is the reason of a unit propagated literal, and if its new activity w.r.t. the current interpretation is better, the activity of $c$ is updated. A clause $c$ is considered better than a clause $c'$ if ${\cal A}(c)< {\cal A}(c')$. This new strategy leads us to the new solver denoted $RelD$-$MiniSAT$.

\begin{table}[h]
\begin{small}
\centering
\begin{tabular}{|l|c|c|}
 \hline
{$Solvers$}  & {\#Solved (\#SAT - \#UNSAT)} & {Average Time}  \\ \hline
%{$Minisat$ $2.2$} &{201 (122 - 79)} & {956.78s}    \\  \hline
{$Glucose$ $3.0$}& {216 (104 - 112)} & {\bf 807.62s} \\  \hline
{$Lingeling$ $2013$}& {{\bf 233} (119 - \bf 114)} & {1090.99s} \\  \hline
\hline 
{$Size(12)D$-$MiniSAT$}  & {{\bf 233} ({\bf 123} - 110)} & {1153.9s} \\  \hline
{$SizeD$-$MiniSAT$}  & {231 (120 - 111)} & {1174.97s} \\  \hline
 \multicolumn{1}{|l|}{$RelD$-$MiniSAT$} & 222 (122 - 100) & {1157.31s}  \\  \hline
%{$Minisat 2.2 + Weight$} & { {\bf 234 (125} - 109)} & {1132.08s}  \\  \hline
\end{tabular}
\vspace{0.2cm}
\caption{A Comparative Evaluation of Dynamic Strategies}
\label{tab:Compar2}
\end{small}
\end{table}

In Table \ref{tab:Compar2}, we present the comparative results of the three new dynamic relevant based deletions strategies. As we can see, $Lingeling$ $2013$ and $Size(12)D$-$MiniSAT$ present comparable performances (233 solved instances). We also observe that $Lingeling$ $2013$ is better (respectively worse) than $Size(12)D$-$MiniSAT$ on unsatisfiable (respectively satisfiable) instances.  The solver MiniSAT integrating our three relevant based deletion strategies  outperforms the solver $Glucose$ $2013$. 

In this section,  three dynamic strategies are proposed  where the activities of the learned clauses are updated dynamically during search. As a summary, size-bounded randomized strategy $SBR(12)$-$MiniSAT$ remains the best, as it solves more instances ($236$ solved instances) than all the proposed strategies proposed in this paper including the state-of-the-art SAT solvers $Lingeling$ $2013$ and $Glucose$ $3.0$. 

All the proposed strategies can be easily  integrated to MiniSAT using few lines of codes.

\subsection{Substituting LBD with  Clause Size Inside Glucose}
In the previous sections, we compared our deletion strategies integrated in MiniSAT with the state-of-the-art SAT solvers $Glucose$ $3.0$ and $Lingeling$ $2013$. These two solvers exploit LBD based measure for managing the learned clauses database. The question that we want to answer is the following: what about substituting LBD measure with clause size inside $Glucose$ $3.0$ and $Lingeling$ $2013$?  As  these two solvers exploit LBD based measure for managing the learned clauses database, we only modify $Glucose$ $3.0$ to evaluate the effect of substituting LBD with dynamic clause size ($Size(k)D$) presented in Section \ref{sec:rbsDyn}. The obtained SAT solver is called $Size(k)D$-$Glucose$ $3.0$, where $k$ is a clause size threshold. This new solver is obtained by substituting LBD with Size in the two following parts of $Glucose$ $3.0$: 
%Let us explain the two modifification  modification is done inside $Glucose$ $3.0$. The only modifications we made are: 
\begin{itemize}
\item {\it Initial learned clause activity}: when a learned clause $c$ is derived, its activity is set as follows: ${\cal A}(c) =|c|$ if $|c|<k$, otherwise ${\cal A}(c) =k+|c|$.
\item {\it Dynamic update of clause activity}: at each conflict, the activity of the learned clauses involved in the derivation of the asserting clause is updated as follows: Let $d$ be the conflict decision level.  if $k+d < {\cal A}(c)$ then ${\cal A}(c) = k+d$. Obviously, the activity of a clause of size less than $k$ is not updated.  
\end{itemize}

The $SBR(k)$-$Glucose$ $3.0$ integrates size-bounded randomized deletion strategy.  It is obtained as follows:  each time a learned clause $c$ is derived, it receives ${\cal A}(c)= |c|$ if $|c|\leq k$, otherwise ${\cal A}(c) = k+irand(random\_seed, |V_{\cal F}|)$, where   $irand(random\_seed, |V_{\cal F}|$) return a number between 0 and $|V_{\cal F}|$. This activity remains unchanged during search. 

\begin{table}[h]
\begin{small}
\centering
\begin{tabular}{|l|c|c|}
 \hline
{$Solvers$}  & {\#Solved (\#SAT - \#UNSAT)} & {Average Time}  \\ \hline
%{$Minisat$ $2.2$} &{201 (122 - 79)} & {956.78s}    \\  \hline
{$Glucose$ $3.0$}& {216 ({104} - 112)} & {\bf 807.62s} \\  \hline
\hline
{$Size(12)D$-$Glucose$}  & {{\bf 224} (103 - {\bf 121})} & {919.16s} \\  \hline
{$Size(15)D$-$Glucose$}  & {219 (100 - {119})} & {944.36s} \\  \hline
{$SBR(15)$-$Glucose$}  & { 222 ({\bf 105} - {117})} & {1001.34s} \\  \hline

\end{tabular}
\vspace{0.2cm}
\caption{A Comparative Evaluation of $Glucose$, $Size(k)D$-$Glucose$ and $SBR(k)$-$Glucose$}
\label{tab:Compar3}
\end{small}
\end{table}

This experiment demonstrates that the clause size is clearly more relevant than LBD/glue measure. As we can see, from the result depicted in Table \ref{tab:Compar3}, {$Size(12)D$-$Glucose$ $3.0$ solves 8 instances more than $Glucose$ $3.0$. More interestingly, on unsatisfiable instances, our version solves 9 instances more than $Glucose$ $3.0$. On satisfiable instances, both solvers present similar behavior. $Glucose$ $3.0$ solves only one additional instance. This experiment give us a definitive illustration that clause size based measure is better than LBD. 

\section{Discussion}
\label{sec:discuss}
As a summary, we demonstrate that the size of the clause remains the best metric to quantify the usefulness of a given clause. The lesson that can be drawn from this study, is that predicting the best clauses to be maintained during search deserve further investigation.  In our opinion, the performance of the LBD measure can be explained by the fact that it is really related to the size of the clauses.  Indeed, we have $2\leq LBD(c)\leq |c|$. The clause size is an upper bound of the LBD measure. For instance, the LBD of binary clauses is $2$.  Hence, the strategy defined in Glucose favors in some sens maintaining short clauses. In \cite{AudemardS09}, the authors mention that "{\it A good learning schema should add explicit links between independent blocks of propagated (or decision) literals. If the solver stays in the same search space, such a clause will probably help reducing the number of next decision levels in the remaining computation}". The intuition given by the authors, to understand the importance of clauses of LBD 2:  "{\it the variable from the last decision level will be glued with the block of literals propagated above. We suspect all those clauses to be very important during search}". Let us discuss these claims. First any learned clause add explicit links between independent blocks of propagated literals. Secondly, any learned clause also help to prune the search space and then to reduce the number of decision. The question we ask is the following: let $c_1 = (x_1\vee x_2\vee x_3)$ and $c_2 = (y_1\vee y_2\vee, \dots, \vee y_n)$ with $n>3$, $LBD(c_1) = 3$ and $LBD(c_2) = 2$, which one is relevant? As you can guess, our preference leans to the first clause. 

%Suppose now, that all the literals of $c_1$ are assigned at levels greater than $1000$ while those of $c_2$ are assigned at levels less than $10$, in this case, such preference becomes less evident. 

%that we have $c$ and $c'$ are of same size, and all the literals are assigned at levels 1 and 2, while all the literals of $c_2$  are assigned at levels greater than $100$, in this case 
%
%they only contain one variable of the last decision level (they are FUIP), and, later, this variable will be ?glued? with the block of literals propagated above, no matter the size of the clause. We suspect all those clauses to be very important during search, and we give them a special name: ?Glue Clauses?.
%
%One needs to collapse independent blocks of propagated literals in order
%to reduce the decision level. 
%Intuitively, it is easy to understand the importance of learnt clauses of LBD 2: they only contain one variable of the last decision level (they are FUIP), and, later, this variable will be ?glued? with the block of literals propagated above, no matter the size of the clause. We suspect all those clauses to be very important during search, and we give them a special name: ?Glue Clauses? (giving the name ?glucose?).

%glue clauses tend to The explanation given by the authors about glue clauses need to be on their ability to create links between different blocks is not really convincing. 

%The best version presented in this paper ($SBR(12)$-$MiniSAT$) is available online at: http://www.cril.univ-artois.fr/$\sim$sais. 
An important remark, all the strategies presented in this paper can be integrated easily (few lines of code) to any CDCL-based SAT solver.   A dedicated web page including the proposed learned clauses database deletion strategies integrated to MiniSAT is currently under construction (http://www.cril.fr/$\sim$sais). 

\section{Conclusion and Futur Works}
In this paper, we addressed the learned clauses database management problem. We demonstrated that size-bounded learning strategies proposed more than fifteenth years ago \cite{Marques-Silva96,Bayardo96acomplexity,Bayardo97} remains a good measure to predict the quality of learned clauses. We have shown that adding randomization to size bounded learning is a nice way to achieve controlled diversification. It allows us to favor short clauses, while maintaining a small fraction of large clauses necessary for deriving resolution proofs on some SAT instances. The experimental results, shows that Size-Bounded Randomized strategy integrated to MiniSAT can achieve better performance than the state-of-the-art SAT solvers. Convinced by the importance of short clauses, we proposed several efficient dynamic variant that aims to maintain short clauses containing literals that occurs more often at the top of the search tree. Our last evaluation shows that substituting LBD with clause size based measure inside Glucose leads to better performance on both SAT and UNSAT instances. This paper, open several perspectives including the refinement of the strategies proposed in this paper. 
%As a future work, we intend to investigate the effects of other related components such as restarts (e.g. \cite{Huang07,PipatsrisawatD09Restart}). 

%and reduction frequency \cite{MiniSat03,AudemardS09}.
%Future works  include the study of the effect of other related components such as restart \cite{Huang07,PipatsrisawatD09Restart} and reduction frequency \cite{MiniSat03,AudemardS09}.
%We also plan to conduct a systematic study of some important deletion strategies (size-bounded and relevance bounded strategies) as described by the authors.  

\bibliographystyle{plain}
\bibliography{biblio,satBib}
\end{document}